\documentclass[10pt, conference, compsocconf]{IEEEtran}
\ifCLASSINFOpdf
  \usepackage[pdftex]{graphicx}
\else
\fi
\usepackage[cmex10]{amsmath}
\usepackage{algorithm}
\usepackage{float}
\usepackage{algorithmic}
\usepackage[algo2e, linesnumbered,ruled,vlined]{algorithm2e}

\usepackage{algorithm2e}

\usepackage{array}

\usepackage{eqparbox}
\usepackage{url}

\begin{document}
%
\title{POPCat: Propagation of particles for complex annotation tasks}


\author{\IEEEauthorblockN{Adam Srebrnjak Yang}
\IEEEauthorblockA{University of Waterloo\\
Waterloo, Canada\\
asyang@uwaterloo.ca}
\and
\IEEEauthorblockN{Dheeraj Khanna}
\IEEEauthorblockA{University of Waterloo\\
Waterloo, Canada\\
d25khann@uwaterloo.ca}
\and
\IEEEauthorblockN{John S. Zelek}
\IEEEauthorblockA{University of Waterloo\\
Waterloo, Canada\\
jzelek@uwaterloo.ca}

}


%


\maketitle


%
\IEEEpeerreviewmaketitle

\begin{abstract}
Novel dataset creation for all multi-object tracking, crowd-counting, and industrial-based videos is arduous and time-consuming when faced with a unique class that densely populates a video sequence. We propose a time efficient method called POPCat that exploits the multi-target and temporal features of video data to produce a semi-supervised pipeline for segmentation or box-based video annotation. The method retains the accuracy level associated with human level annotation while generating a large volume of semi-supervised annotations for greater generalization.
The method capitalizes on temporal features through the use of a particle tracker to expand the domain of human-provided target points. This is done through the use of a particle tracker to reassociate the initial points to a set of images that follow the labeled frame. A YOLO model is then trained with this generated data, and then rapidly infers on the target video. Evaluations are conducted on  GMOT-40, AnimalTrack, and Visdrone-2019 benchmarks. These multi-target video tracking/detection sets contain multiple similar-looking targets, camera movements, and other features that would commonly be seen in "wild" situations. We specifically choose these difficult datasets to demonstrate the efficacy of the pipeline and for comparison purposes. The method applied on GMOT-40, AnimalTrack, and Visdrone shows a margin of improvement on recall/mAP50/mAP over the best results by a value of 24.5\%/9.6\%/4.8\%, -/43.1\%/27.8\%, and 7.5\%/9.4\%/7.5\% where metrics were collected.
  \end{abstract}

\begin{figure*}[h]
    \centering
    \includegraphics[width=1\linewidth]{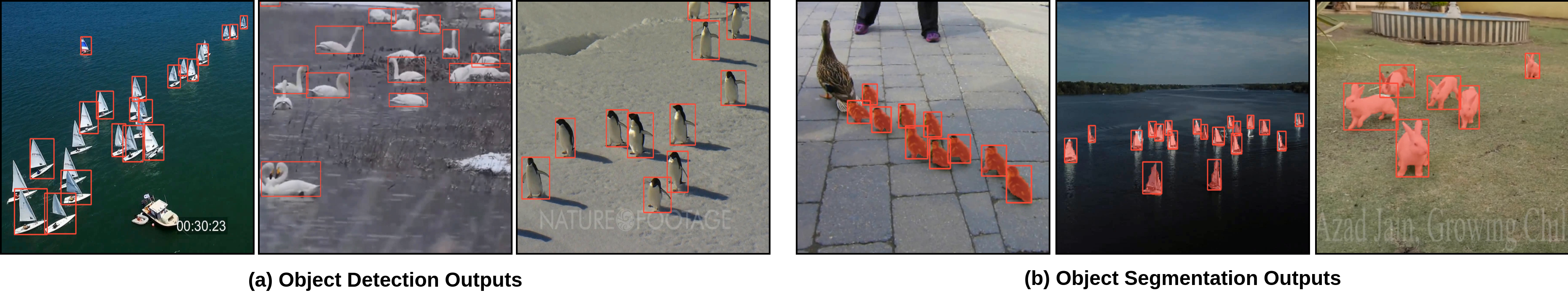}
    \caption{Detection and Segmentation Outputs on GMOT-40\cite{bai2021gmot} and AnimalTrack\cite{animaltrack}}
    \label{fig:enter-label}
\end{figure*}
\section{Introduction}
\label{sec:intro}
Most learning-based computer vision tasks and algorithms utilize a labeled dataset to complete a desired task. As such, we think it prudent to develop a faster, scalable, and efficient way to annotate datasets. We focus on multi-target video dataset creation as it forms the basis for the popular tasks of crowd counting and multi-object detection. Video datasets such as KITTI\cite{Geiger2013IJRR}, MOT\cite{MOTS20}, and GMOT-40\cite{bai2021gmot} utilize image-based annotation methodologies to create the respective datasets. These methods individually annotate all frames of the video sequence with human-derived methods. We improve the process by utilizing a trained model to semi-automatically annotate all targets within the video. This allows for a semi-supervised approach for dataset annotation.

We exploit the multi-target and temporal features of videos to produce a semi-supervised pipeline for segmentation or box-based video labeling. We enhance the process by utilizing computer vision methods to train a detector for rapid inferencing. This, therefore, augments our method with high speed labeling once the algorithm is trained on a sparse set of human annotations.

We separate our pipeline into four stages: (a) initialization, (b) propagation, (c) segmentation and box fitting, and (d) model training. We detail the operation of each stage under Section \ref{sec:pipeline} and depict it in Fig. \ref{fig:pipeline}.

The input to the pipeline is an annotated frame with the output being the detected per-frame labels for the entire video sequence. We state that this generic, multi-object, video annotating solution allows for the generation of per-use-case datasets for computer vision algorithms. We designate the pipeline as "POPCat" as it utilizes the propagation of particles to address complex annotation tasks such as multi-target video datasets.

Modern object detectors adopt an out-of-the-box application with minimal modification required for implementation (YOLO\cite{Yolo}, R-CNN\cite{RCNN}, and DETR\cite{carion2020end}). We adopt the YOLO detector series in our pipeline as it has long-term support and is readily used in industrial computer vision. This familiarity makes the adoption of our pipeline an easy process, with users already familiar with the main components.

To be direct, the intended users for POPCat are the smaller-scale computer vision operations with lesser access to annotators, compute power, or cloud resources. We keep this in mind throughout development which enforces restrictions on the computational requirements of the system. For example, a transformer-based architecture may be advantageous for recall and real-time for propagation, but it may be too computationally expensive for the target audience. Most industrial setups are limited to low power compute units which impose a VRAM limitation on the system. A balance has to be struck between computational requirements and pipeline performance in order to reach a greater target audience. We artificially set the graphics requirement for our system at 12GB of VRAM to broaden the potential user base. 

The pipeline excels in situations where multiple targets are present on a single frame as it can scale up the total labels by the number of propagated frames. For example, a 40-point target on the initial frame can lead to 1200 labeled instances via label propagation over 30 frames without the use of the integrated detector. Examples of the segmentation and detection outputs are shown in Fig. \ref{fig:enter-label}. 

The contributions of our paper can be separated into four distinct sections:
\begin{enumerate}
    \item The development of a semi-supervised annotation pipeline based on off-the-shelf components.
    \item The use of Segment Anything (SAM)\cite{kirillov2023segany} for automatic bounding box resizing.
    \item The training and application of detection modules for rapid labeling of contextually similar video frames.
    \item Benchmarking of our detections against ground truths provided in the GMOT-40\cite{bai2021gmot} dataset, AnimalTrack dataset\cite{animaltrack}, and VisDrone-2019\cite{9021972} Video detection dataset. 
\end{enumerate}

\section{Related Work}

\label{sec:related-work}

\subsection{Industrial Applications}
\label{subsec:indust-app}
Our team reviews numerous industrial implementations of computer vision systems to identify the common characteristics of industrial settings and systems. A common focus is kept on the efficiency of computer vision systems as human-hours are held at a higher value than computational loading\cite{MOSP,rutinowski2023semi,rasmussen2022challenge, dorr2020fully,PALLETFORK,9613245,COAL}. We divide these works into manually annotated training sets \cite{dorr2020fully,COAL,9613245,rasmussen2022challenge, PALLETFORK,9204369}, and semi-automated annotation \cite{MOSP,rutinowski2023semi}. 

Multiple authors note that manual annotations may not effectively cover all situations in production environments \cite{COAL,dorr2020fully,PALLETFORK,9613245} and additional training data may be required. These systems represent the brute force method for industrial vision systems where the end user is required to manually annotate a training set. Such datasets may fail to properly generalize when the background or context of the targets changes. This may require a new dataset to be generated to retrain the algorithm if the operational environment changed\cite{COAL,dorr2020fully}.

The semi-automated training methods utilize color space thresholds \cite{MOSP}, and motion capture \cite{rutinowski2023semi} to recognize industrial targets. The color space threshold method is used to identify cardboard boxes from an overhead view\cite{MOSP}. This method benefits from good contrast between target and background, and sparse number of targets on the screen\cite{MOSP}. We note that threshold-based methods would struggle within dense multi-target environments as there is no way to distinguish overlapping parts from one another. Furthermore, it requires good contrast between the background and target which cannot be guaranteed in production. The motion capture system of \cite{rutinowski2023semi} utilizes a series of RGB cameras to collect images and fuse the data with the motion capture system. The pose extracted from the motion capture system is overlayed on different camera views to generate a significant amount of data. Annotation is required on the motion capture data, and is then propagated to the different camera views. The system is capable of annotating tens of thousands of frames with an average annotation time of 0.2 frames per second for a multi-target setup. However, the complexity of the system is too high for most industrial setups. We understood that an effective system must be able to annotate a significant number of frames with minimal input as shown in \cite{MOSP,rutinowski2023semi}, however, a production system must do it with minimal installation requirements. 

Our review of current industrial implementations highlights three facts about datasets. Namely, datasets are an integral part of training the vision system, must be related to the detection targets, and are expensive to create.
\subsection{Zero-Shot Object Detection}
\label{subsec:one-shot-few-shot}
In addition to the methods of industrial implementations, many computer vision systems rely on zero-shot learning to overcome the requirement for large custom datasets. These methods have grown rapidly in previous years with the introduction of large language transformer models such as BERT\cite{devlin2018bert}, and GPT\cite{Radford2018ImprovingLU}. We limit our discussion here to Z-GMOT\cite{tran2023z} as it is relevant to the later evaluations. Semantic object detection utilize foundational models to identify targets based on previously seen data\cite{tran2023z}. Algorithms such as Z-GMOT\cite{tran2023z} utilize a grounded language inference process (GLIP)\cite{li2022grounded} to semantically search for target objects within images. Further textual descriptors are applied to enhance the accuracy of the detections\cite{nguyen2022few,tran2023z}. 
Our team noted that the quality of the output detection may differ depending on the prompts provided to the semantic detector\cite{tran2023z}. Furthermore, novel classes that are not within the model must be trained or learned by the model\cite{SEMSEG}. At this time, semantic models are deemed too complex and potentially computationally expensive for use within industrial dataset annotation/creation. 
\begin{figure*}[htbp]
  \centering
\includegraphics[width=17.5cm]{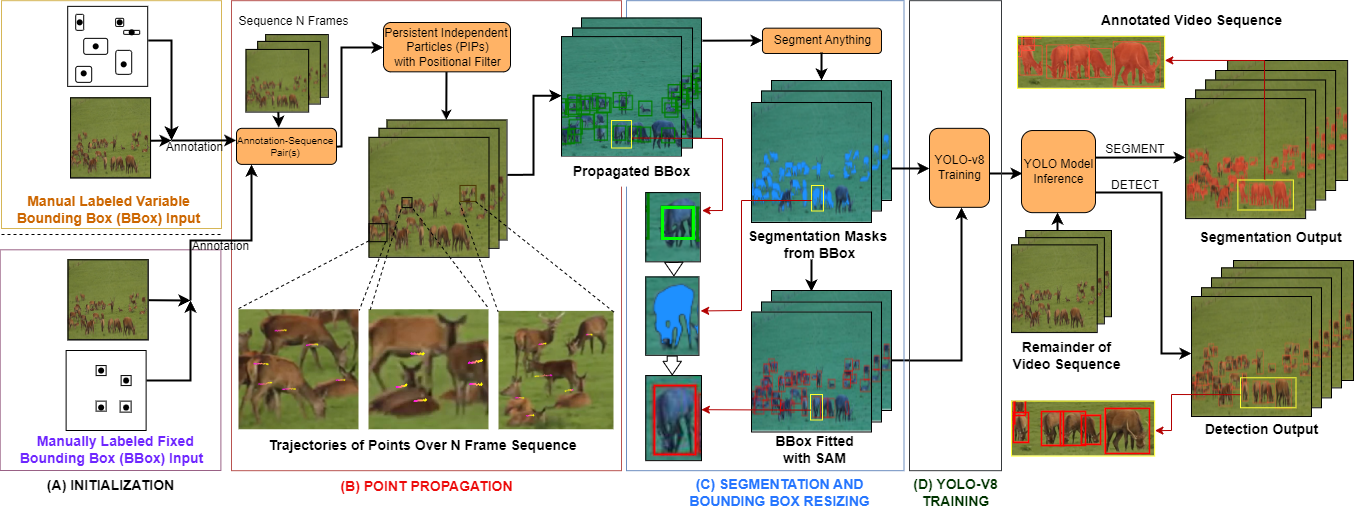}
  \caption{POPCat Pipeline Representation}
  \label{fig:pipeline}
\end{figure*}

\section{Methodology} 
\label{sec:pipeline}
The purpose of this section is to describe each stage of operation of the pipeline. We cover the target domain, and follow through with the initialization, point propagation, segmentation and box fitting, model training, and model application. The pipeline is depicted in Fig. \ref{fig:pipeline}

\subsection{Target Domain} 
\label{subsec:targetdata}
The target domain for our system is industrial vision systems as the environment synergizes with the proposed system. Industrial environments tend to feature a statically mounted camera or view with a fixed context, that contains numerous targets moving along a fixed path. This case has been identified during the literature review, and supports the methodology used in training. Namely, multiple targets can be annotated on one given frame, and the path of travel can be learned through the point propagation methodology. Once learned, a detector can be used to rapidly annotate the remainder of the video.

\subsection{Initialization} 
\label{subsec:instantiation}
We initialize the POPCat pipeline with a manual frame annotation as seen in (A) of Fig. \ref{fig:pipeline}. The manual annotation is supplied with a sequence of subsequent frames to create an annotation-sequence pair. 

For a 1200-frame video, a single first-frame annotation for a 30-frame portion of the video would represent an annotation-sequence pair. We reduce the total number of necessary manual annotations to the total number of annotation-sequence pairs used. For most video sets, a single annotation-sequence pair is required. 

Frame initiation can take one of two forms: (a) variable box selection and (b) fixed box selection. \textbf{Variable box selection} requires a fitted bounding box for each target and is applied when targets take on a variety of sizes over the provided sequence. These variations may occur due to viewing angle, foreground-background travel, rapid target movement, and other variables. Variable box selection requires more annotation time as an individual must manually mark the extent of target boxes within the first frame image. Conventional online annotation tools such as CVAT\cite{CVAT} can be used to expedite the first frame labeling process for variable sized selection. We represent variable box selection in Algorithm \ref{alg:pipeline_alg} lines 1-2.

The latter labeling method is \textbf{fixed box selection} which is considered the faster of the two methods. In this method, a user sets a fixed box width and height, and must then click closest to all of the center points of the targets to be tracked. This method prioritizes speed as the user only sets the box sizes once, and then clicks on the center of targets. Based on this setup, a single annotator can rapidly annotate a given image by selecting the center point of targets within the image. The fixed box size should fully encompass the majority of the targets marked for propagation. However, overtly large boxes are not recommended as overlapped parts may be segmented as one large part. Therefore, for scenarios where the target takes on a multitude of shapes, it is recommended that the dimensional average of all boxes be used for propagation. We represent fixed box selection in Algorithm \ref{alg:pipeline_alg} lines 4-5.

The precision of the labels is important as the point must be present on the intended targets for tracking. If the points are not on target then the tracker may track the background movement instead. To alleviate this we upscale small scale images and allow the user to zoom in on targets to ensure correct selections. Once labeled, the annotations are passed to the point propagation stage.

\begin{algorithm}[H]
\caption{Label propagation \& Segmentation Algorithm (STAGES A-C)}
\label{alg:pipeline_alg}
 \KwIn{$K,M$: No. of points/bbox in a frame, $M<=K$\\
 \hspace{1cm} $N$: No. of frames propagated \\
 \hspace{1cm} $P_{(j)}$: Set of points in $j^{th}$ frame\\
 \hspace{1cm} $P_{f(j)}$: Updated $P_{(j)}$ points after positional filtering
 }
 \KwOut{$Mask_l$: Segmentation masks for $N$ frames\\
 \hspace{1.2cm} $BBox_l$: Yolo format compatible BBox annotation\\
 \hspace{2.6cm}for $N$ frames
 }
 \tcc{Annotation initialization}
 \If{variable BBox}
    {
        $P_{(1)}^K = (x_i, y_i, w_i, h_i) \forall i \in \{1...K\}$}
 \Else{
 
    \If{center points}
    {
        $P_{(1)}^K = (x_i, y_i, W, H) \forall i \in \{1...K\}$
    }
 }
 \tcc{Point tracking and positional filtering}
 \For{frames j =1 \KwTo N-1}{
    $P_{(j+1)}^K \gets PIPs(P_{(j)}^K)$
  
    $P_{f_{(j+1)}}^M \gets PositionalFiltering(P_{(j+1)}^K)$
  
  }
\tcc{Segmentation and Bounding Box Resizing}
 
\For{frames $l$ =1 \KwTo N}{
  $Mask_l \gets SegmentAnything(P_{f_{(j)}}^M)$
  
  $Mask_l \gets CannyEdgeDetection(Mask_l)$
  
  $BBox_l \gets GetMinBoundingBox(Mask_l)$
} 
\Return $(Mask,BBox)$

\end{algorithm}

\subsection{Point Propagation} 
\label{subsubsec:point-prop}
Persistent Independent particles (PIPs)\cite{harley2022particle} is used for point propagation for the proposed pipeline \cite{harley2022particle}. This method effectively tracks the center points of the marked targets as a particle within the video. We choose not to track the targets as a bounding box as it would be computationally inefficient and less robust to changes in background and box sizing. PIPs utilizes a fully temporal approach to particle tracking which results in highly accurate point tracks across the propagated frames. The points are tracked in 8 frame sequences to overcome momentary occlusions in the video \cite{harley2022particle}. However, due to a purely temporal focus, false positive tracks can occur when points travel out of view.

\textbf{Positional Filtering}: Point tracking in PIPs does not define the termination of any tracked point if the tracked object leaves the field of view. PIPs re-initialize on the 8th frame, therefore a lost particle would re-initialize to the last known position of the particle. In the abscence of occlusions, this manifests as point tracking of the border of the image after a target leaves the frame of view. Therefore, we apply a positional filter to terminate the tracks once the target reaches the edge of the field of view. This eliminates track drift in highly mobile targets. 

The input to PIPs are the center point and box dimensions of the labels marked in the initialization. We propagate the center points through PIPs and keep the size of the bounding boxes for the tracked point the same to modularize this stage. We apply the same operations in PIPs regardless of initialization choice as a way of simplifying the design and configuration of the pipeline for the end user.

The output of this stage is a set of text files for each frame, where each file contains the normalized YOLOv8 data representation of the propagated and filtered targets. These populated text files are fed into the segmentation stage of the pipeline. We represent the use of PIPs and the positional filter in Algorithm \ref{alg:pipeline_alg} lines 6-8.

%
%
%
%
%
%
%
%
\subsection{Segmentation and Bounding Box Resizing} 
\label{subsec:segment-bbox-resizing}

\begin{figure}[h]
\includegraphics[width=8cm]{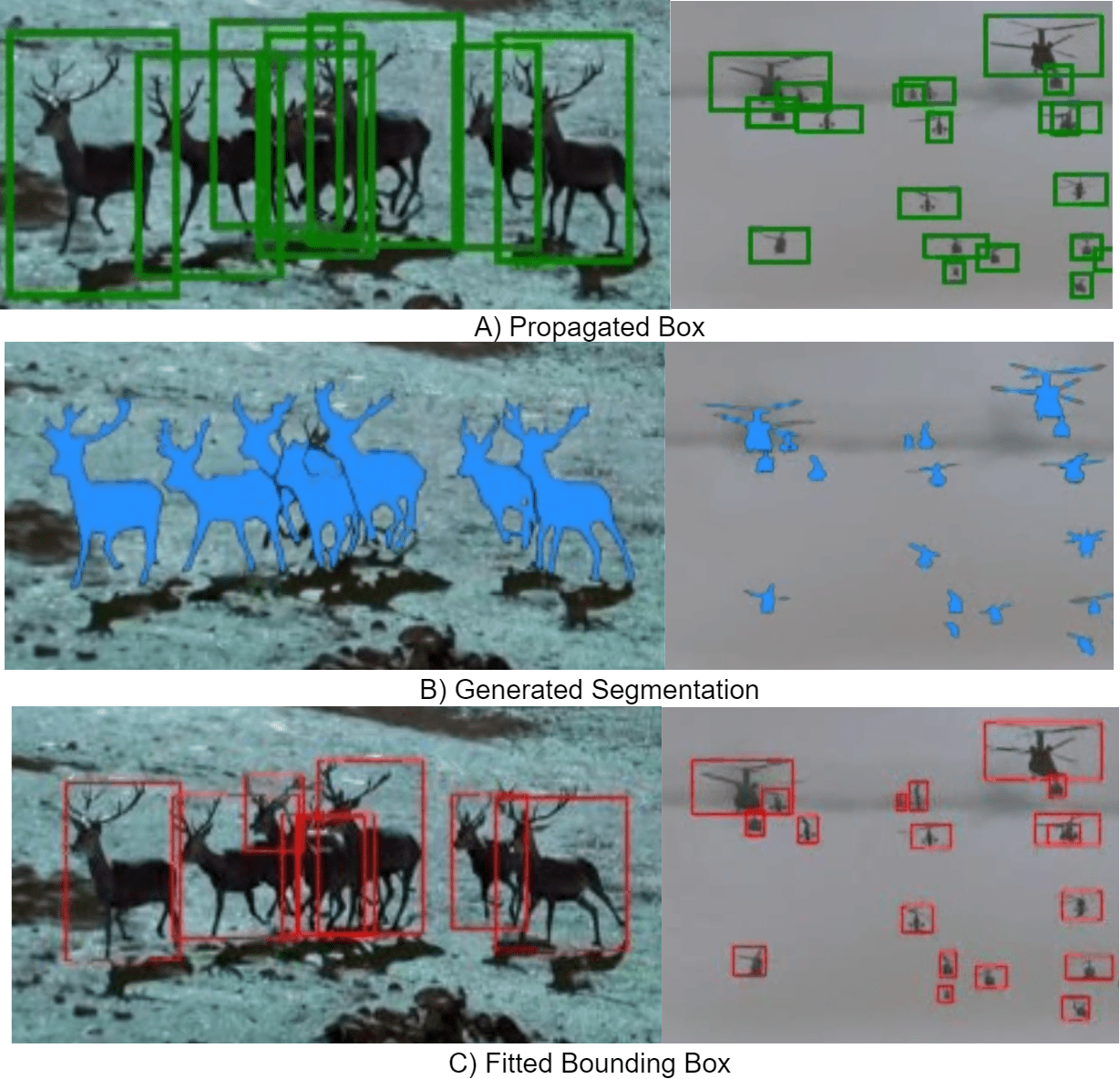}
    \centering

 \caption{Application of Segment Anything Model within the POPCat pipeline}
 \label{fig:segmentfit}
\end{figure}

We apply a segmentation module to enhance the accuracy of box annotations centered on the propagated points. We utilize Segment anything Model (SAM) as it is a zero-shot, prompt based segmentation model that is capable of segmenting most well defined shapes\cite{kirillov2023segany}. We utilize the propagated boxes from PIPs (center point with height and width) as the prompts for automatic segmentation of targets within each frame. We depict this in Fig. \ref{fig:segmentfit}. A careful review of the failure cases of SAM in \cite{ji2023segment} shows no failure modes that adversely affect the intended use case. A box is applied around the target to eliminate much of the over-or-under segmentation issues associated with SAM \cite{ji2023segment}. Given that SAM is a foundational model, it requires no further training to effect accurate segmentation masks. 

We utilize SAM to effect a process known as \textbf{box resizing}. Works such as Grabcut \cite{grabcut} and Autofit \cite{autofit} demonstrate how precision and recall increased proportionately to improvements in the bounding box fit. We demonstrate this process in Fig. \ref{fig:segmentfit} where a) shows the originally labeled boxes and c) shows the ones generated via the segmentation process. To generate these boxes we apply a canny edge detector on the segmentation mask to extract the contours of the segmentation. From there, a minimum bounding rectangle is applied to the contours to effect the minimum box representation of the mask. The contours are also saved for training of the segmentation detector in subsequent stages. One should note in Fig. \ref{fig:segmentfit} (a) to (c) that the box resizing method can increase and decrease the size of bounding boxes to capture the full content of the target. 

The application of SAM as shown in Fig \ref{fig:segmentfit} is represented in Algorithm \ref{alg:pipeline_alg} lines 9-12. This includes the extraction of segmentation masks and the creation of fitted bounding boxes. The output of this stage are YOLO formatted text files that include bounding box dimensions in one set of text file and segmentation x,y tuples in another set of files.

\subsection{Training of Yolo Model} 
The goal of the previous stages was to propagate a first-frame annotation across a sequence to increase the number of annotated instances in an automated manner. Our pipeline now diverges from most mainstream methods as we utilize the generated instances to train a detector or segmentation YOLO-v8 model as depicted in (D) of Fig. \ref{fig:pipeline}. The pipeline provides the training data necessary to effect either a box or segmentation model depending on the annotation requirements. We note that any detector could be used in place of the YOLO model as our inputs to this stage is just training data. This modularity is intentional as it allows for the adaptation of the pipeline for more specialized use cases. The details of training will be covered under the implementation section \ref{subsec:impl-details}. 

\subsection{Application of Trained Models} 

By training an object detector based on the instances generated with Algorithm \ref{alg:pipeline_alg}, we leverage the speed of the YOLO detection algorithm for high frame volumes. The numerical advantage is shown as follows. A 10-minute at 30fps video contains 18000 frames. Assuming a 20 point first frame annotation for point propagation, Tab. \ref{table:time-breakdown} showcases the breakdown of time taken in \textit{seconds} to process each stage of our pipeline. The example demonstrates the effectiveness of our solution in the areas of human cost (1 labeled frame), and time efficiency (10 FPS). For comparison purposes, a 1 FPS rate is retrieved from the Youtube-BB dataset\cite{8100272}, which focuses on the manual annotation of multi-target videos\cite{8100272}. This dataset uses a system of expert annotators to quickly label multi-target videos in the wild. We note that our system has a 10:1 advantage in speed, and only requires the use of a single expert annotator. We do not evaluate other detectors by training them on our generated annotations as it is not a comparable training methodology. That is, we would effectively be comparing a system trained on all frames to one that was trained on a subset for a challenge. 

\begin{table}[htbp]
\centering
\begin{tabular}{p{6cm}|cc} 
\hline
    \textbf{Stage} & \textbf{Time (sec)} \\ \hline 
    \textit{First frame labeling} & 120  \\
    \textit{PIPs Propagation} & 400  \\
    \textit{SAM Propagation} & 300  \\ 
    \textit{YOLO Model Training} & 360  \\ \hline
    \textit{Yolo-Inference(remaining frames)} & ~600 \\ \hline
    \textbf{Ours Total time to annotate} & \textbf{1780} $\sim$ \textbf{10FPS} \\ \hline
    \text{Youtube-BB multi-annotator manual method} \cite{8100272} & 1 FPS \\ 
\hline
         
\end{tabular}

\caption{Time Breakdown for annotation of a 10 minute video}
\label{table:time-breakdown}
\end{table}
\section{Experiments and Results}
\label{sec:Exp}
\subsection{Datasets}
The Visdrone-2019 video detection challenge \cite{9021972}, GMOT-40\cite{bai2021gmot}, AnimalTrack\cite{animaltrack} sets are used to benchmark the performance of POPCat. These sets contain high-quality ground truth bounding box data for video sequences of varying duration. Set variations in video data included lighting, object sizing, movement, occlusion, camera position, camera motion, and target volumes.

The Visdrone-2019 challenge \cite{9021972} test-dev set contains multi-class detection data, variable-sized classes, and variations in video sizing. Our method utilizes three annotation-sequence pairs from the start, middle, and end of each video. This is done to address temporal class imbalances in the video. Temporal class imbalance refers to the situation where classes at the start of a sequence are not representative of the end of the sequence. This is simply a heuristic method to address wide variations in the context of the video. There is still the possibility that certain classes will be missed through this method, however, we want to benchmark the effectiveness of the system with limited data. We train a single detector model for validation on all videos in the test-dev set to match the challenge protocol. We note that the results published by the Visdrone team refer to the challenge dataset of the competition. That dataset differs from the test-dev set which is publicly available. Subsequent works benchmark results on the test-dev set \cite{9573394,Fujitake_2022_WACV} and are referenced for comparison.

GMOT-40\cite{bai2021gmot} and AnimalTrack\cite{animaltrack} are tracking datasets that benchmark detection protocols with a variety of trackers. GMOT-40 and AnimalTrack feature numerous targets per frame in a single-class configuration. We train on each video separately in our pipeline with a single start frame annotation-sequence pair. Our training methodology matches the GMOT-40 one-shot protocol to allow for results comparison. AnimalTrack detectors utilize a subset of dataset for testing, and our reported results are collected on the same subset. Both ablation studies are conducted on the full sets of GMOT-40 and AnimalTrack.

\subsection{Implementation Details}
\label{subsec:impl-details}
Starting with point propagation, we develop a modified variant of the chain-demo.py implementation from PIPs \cite{harley2022particle}. This implementation features a dictionary which keeps track of the tracking id, box dimensions, and position of the points being propagated. A positional filtering subroutine is implemented within this section to eliminate stray points. We set the resolution at 1280x720 with a 30 frame sequence. This is done to adhere to the VRAM constraints placed on the system. 

Next, SAM is modified to operate in a batched sequence with an image resolution of 1280x720. The resolution is upscaled to effect accurate segmentations in operation. We predict in batches of 50 points at a time and concatenate the results at the end in order to limit the amount of GPU ram usage. This allows the system to operate on older or more constrained hardware setups.

Our solution uses a fixed training and validation confidence of 0.2, with an IOU of 0.5 with agnostic\_nms enabled for the yolox.pt weight configuration for the detector model. Training epochs are fixed at 25 for all YOLO models, with a batch size of 12, image size of 640, randomized model weight, SGD optimizer, learning rate of $0.01$, and default parameters as dictated by YOLO for all processes\cite{Yolo}. Data augmentation methods are disabled during training. Propagation, segmentation, model training, and inference are performed on a single NVIDIA Titan X GPU. 

With this implementation, the pipeline is able to maintain the low gpu compute resource constraints while annotating a significant amount of data.

\subsection{Evaluation Metrics}
The results are compared to other methods through mean average precision (mAP), mean average precision at 50 percent intersection over union (mAP50), and recall when available. This information details the effectiveness of our algorithm when compared to the ground truth data. For example, a recall of 75\% is indicative that our system is capable of noting 75\% of the total labels based on the applied pipeline. For data referenced in Table \ref{table:ANIMGMOT-40-results}, recall and precision was available on a case by case basis. For a more detailed analysis, we state precision, recall, true positive (TP), false positive (FP), false negative (FN), mAP50, mAP, and F1 confidence scores for the pipeline. This should allow future comparison of our solution with one-shot, few-shot, and semantic-based methods. It is noted that a proper annotation method should minimize false positives while maximizing true positives to enhance downstream detector performance\cite{ristani2016performance,electronics10030279}.

\begin{table}[htbp]
\centering  
\scalebox{0.9}{
\begin{tabular}{p{4.5cm}|ccc}
\hline
\textbf{GMOT-40 - Method} &  \textbf{Recall\%} & \textbf{mAP50\%} & \textbf{mAP\%} \\ \hline
    \textit{GlobalTrack \cite{GlobTrack}GMOT-40\cite{bai2021gmot}}  & - & 15.65 & - \\ 
    \textit{Z-GMOT GLIP\cite{tran2023z}}  & - & 66.20 & 36.10  \\ 
    \textit{Z-GMOT iGLIP\cite{tran2023z}}  & - & 66.90 & 40.00 \\ 
    \textit{Siamese-DETR (COCO\cite{10.1007/978-3-319-10602-1_48})\cite{liu2023siamesedetr}}  & 49.90 & 63.60 &-  \\ 
    \textit{Siamese-DETR (Objects365\cite{Shao_2019_ICCV})\cite{liu2023siamesedetr}}  & 55.40 &69.60 & -\\ \hline
    \textit{Fixed Box SAM Pos.(Ours)}  & 74.97 & 74.92 & 40.62 \\
    \textit{Variable Box SAM Pos.(Ours)}  & \textbf{79.93} & \textbf{79.19} & \textbf{44.81}  \\ \hline 
    \textbf{AnimalTrack Subset - Method} &  \textbf{Recall} & \textbf{mAP50} & \textbf{mAP}\\  \hline
    \textit{Faster-RCNN \cite{ren2016faster} AnimalTrack\cite{animaltrack}} & - & 34.40 & 16.10  \\
\hline
    \textit{Fixed Box SAM Pos.(Ours)}  & 64.04 &70.57 & 35.61  \\
    \textit{Variable Box SAM Pos.(Ours)}  & \textbf{72.55} &\textbf{77.57} & \textbf{43.90}  \\
\hline
\end{tabular}}

\caption{GMOT-40 and AnimalTrack Subset Detection Comparison. Comparison results obtained from \cite{bai2021gmot,animaltrack,tran2023z,liu2023siamesedetr}. (-) means no metric provided by publication. Fixed/Variable Box indicates provided box format, SAM indicates presence of SAM module, Pos. indicates presence of positional filter. Best results are in \textbf{bold}.}
\label{table:ANIMGMOT-40-results}
\end{table}

\begin{table}[htbp]
\centering  
\scalebox{0.9}{
\begin{tabular}{p{4.5cm}|ccc}
\hline
\textbf{Method} &  \textbf{Recall\%} & \textbf{mAP50\%} & \textbf{mAP\%} \\ \hline
    \textit{Faster R-CNN\cite{ren2016faster}\cite{9573394}}  & 13.55 & 26.83 & 10.25  \\
    \textit{CornerNet\cite{law2020cornernet}\cite{9573394}}  & 24.03 & 28.37 & 12.29  \\
    \textit{CenterNet\cite{9010985}\cite{9573394}}  & 24.87 & 28.93& 12.35  \\
    \textit{FPN\cite{Lin_2017_CVPR}\cite{9573394}}  &25.59 & 29.88 & 12.93  \\
    \textit{D\&T\cite{feichtenhofer2017detect}\cite{9573394}}  & 25.64 & 32.28 & 14.21  \\
    \textit{FGFA\cite{zhu2017flow}\cite{9573394}}  & 27.21 & 33.34 & 14.44  \\\hline
    \textit{FCOS\cite{tian2019fcos} Baseline WACV22' \cite{Fujitake_2022_WACV}}  & 26.98& 32.42& 11.44 \\
    \textit{Video-Rep. WACV22' \cite{Fujitake_2022_WACV}}  & 41.28 & 49.01 & 21.82 \\\hline
    \textit{YoloV8 Trained Control\cite{yolodat}}  & 46.50& 49.50 & 27.50  \\ \hline
    \textit{Multi-Class Variable SAM Pos.(Ours)}  & \textbf{48.80} & \textbf{58.40}& \textbf{29.30}  \\ 
    \textit{Single-Class Variable SAM Pos.(Ours)}  & \textbf{60.30}& \textbf{72.00} & \textbf{37.70}   \\  \hline
\end{tabular}}
\caption{Visdrone-2019 Task 2 Video Detection Results on test-dev set. Our trained detector is validated against results provided by \cite{9573394,Fujitake_2022_WACV}. YOLOv8 trained control model is used for comparison purposes\cite{yolodat}. Best results for the comparion are in \textbf{bold}.} 
\label{table:visdrone-results}
\end{table}
\subsection{Discussion}
Multi-class and single-class performance are explained separately. Starting with single class datasets, Tab. \ref{table:ANIMGMOT-40-results} shows that the pipeline outperforms all available works under the reported recall, mAP50, and mAP metrics.
The proposed solution performs the best by a margin of 19.5\%, 5.3\%, and 0.6\% under the fixed regiment of GMOT-40 for recall, mAP50, and mAP. The variable regiment outperforms the competitors by a margin of 24.5\%, 9.5\%, and 4.8\% to the corresponding metric leader. 


Our pipeline performance over the test set of AnimalTrack\cite{animaltrack} is reported in Tab.\ref{table:ANIMGMOT-40-results}. POPCat performs 36.1\%, and 19.5\% better in mAP50 and mAP results when compared to the faster-RCNN\cite{ren2016faster} baseline for fixed selection. Results are averaged to a 43.1\%, and 27.8\% improvement in mAP and mAP50 for the variable selection cases. 

Benchmarking on the Visdrone-2019 test-dev dataset shows that our pipeline outperforms all other detectors by a minimum margin of 21.6\% in recall, 25.0\% in mAP50, and 14.8\% in mAP Tab. \ref{table:ANIMGMOT-40-results}\cite{9573394,Fujitake_2022_WACV}. We acknowledge that our system has access to 54 frames out of the total 6635 frames of the testing dataset. Our pipeline is able to capture context specific data about the environment and targets which leads to significantly better results. We argue that production applications would allow for a calibration phase for vision systems which allows for training of the system on the intended operational views. To validate the operation of yolov8 we utilized a pretrained yolov8-x.pt model from huggingface \cite{yolodat} and validated it with the same setting on Visdrone dataset. Our pipeline outperforms the control test by 2.3\% recall, 8.9\% mAP50, and 1.8\% mAP. This test demonstrates that the labeling methodology has a significant impact on the performance of the yolov8 detector. Furthermore, it demonstrates that the YOLOv8 architecture has an excellent capacity to learn and adapt to difficult datasets.

\begin{figure}[h]
\includegraphics[width=8cm]{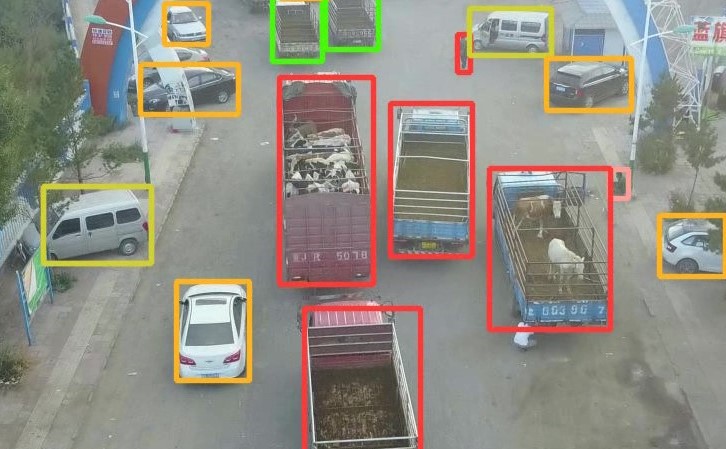}
\centering
 \caption{Class switching is seen in multi-class inferencing on the Visdrone-2019 dataset. Video uav0000077\_00720\_v shows pedestrians as boxed in red, cars boxed in orange, trucks in green, people sitting in pink, and vans in yellow. }
 \label{fig:classswitch}
\end{figure}

A third experiment quantifies the effect of class mismatches in the model. Class mismatch in the multi-class model refers to instances where the boundaries of an object are marked with the incorrect class. We depict class mismatch in Fig.\ref{fig:classswitch} where the trucks are misclassified as pedestrians (red). Under the Visdrone challenge, class mismatch occurs as a result of class imbalance and the sparse labeling methodology applied. To verify this phenomenon, we perform a single-class experiment on Visdrone and evaluate the precision and recall of the system. The single class model shows an improvement of 11.5\% recall, 13.6\% mAP50, and 8.4\% in mAP Tab.\ref{table:visdrone-results}. This means that the detector is still capable of detecting all listed targets but struggles with class assignment. We postulate that including a classifier stage at the end of the system may improve multi-class results. We note that there is still room for improvement within the propagation module as shown in this experiment.

With this pipeline, a total of 149 ground truth annotations (40 GMOT-40, 58 AnimalTrack, 51 Visdrone) generate 4470 labeled frames for 115 separate videos (40 GMOT-40, 58 AnimalTrack, 17 Visdrone). We train 99 detection models to inference a total of 40947 frames (9603 GMOT-40, 24709 AnimalTrack, 6635 Visdrone 2019 Test-set dev) with the data above. The manual frame-to-inference frames ratio is 1:274, making our pipeline extremely efficient for video annotation. Furthermore, our work exceeds all previous works under the fixed box selection with SAM and positional filtering, and has further improvements with the variable regiment. We believe the best use case for our pipeline are industrial computer vision settings. These environments tend to feature multiple targets, a fixed camera, and well-defined backgrounds which synergize with our pipeline.

\begin{table*}[htbp]
\centering 
\label{table:ablation_study}

\resizebox{\textwidth}{!}{%
\begin{tabular}{p{8cm}|cccccccccc}
\hline
\textbf{GMOT-40}\cite{bai2021gmot} - \textbf{Method - Total Instances:255235} &  \textbf{Precision}$\uparrow$ &  \textbf{Recall\%}$\uparrow$ &  \textbf{TP}$\uparrow$ &  \textbf{FP}$\downarrow$ &  \textbf{FP\%}$\downarrow$ & \textbf{FN}$\downarrow$ &\textbf{FN\%}$\downarrow$ & \textbf{mAP50}$\uparrow$ &\textbf{mAP}$\uparrow$  & \textbf{F1}$\uparrow$ \\ \hline
    \textit{Fixed Box Selection / No SAM / No Positional Filter}  & 71.54 & 61.33 & 156536 & 62269  & 24.40 &98699  & 38.67 &58.63&28.04 &66.04 \\ 
    \textit{Fixed Box Selection / No SAM / Positional Filter}  & 72.81 & 61.70& 157485 & 58815 &23.04 & 97750  &38.30 &60.33&27.78 &66.80 \\ 
    \textit{Variable Box Selection / No SAM / No Positional Filter}  & 88.43 & 74.38 & 189841 & 24831 &9.73 & 65394  &25.62 &77.35& \textbf{46.02}&80.80 \\ 
    \textit{Variable Box Selection / No SAM / Positional Filter}  & 89.36 & 74.68 & 190599 & 22687 &8.89 &64636 & 25.32 &77.65&45.17 &81.36 \\ 
    \textit{Fixed Box Selection / SAM / No Positional Filter}  & 86.68 & 74.70 & 190657 & 29312  &11.48 &64578 &25.30 &74.58 &40.10 &80.24 \\ 
    \textit{Fixed Box Selection / SAM / Positional Filter}  & 86.98 & 74.97& 191336 & 28651 &11.22 &63899 &25.03 &74.92&40.62 &80.53 \\ 
    \textit{Variable Box Selection / SAM / No Positional Filter}  & 90.21 & 78.87 & 201307 & 21853 &8.56 &53928 &21.13 &78.91&44.67 &84.16 \\ 
    \textit{Variable Box Selection / SAM / Positional Filter}  & \textbf{90.59} & \textbf{79.93} & \textbf{203998} & \textbf{21196} &\textbf{8.30} &\textbf{51237} &\textbf{20.07} &\textbf{79.19} &44.81 &\textbf{84.92} \\ \hline
    \textbf{AnimalTrack}\cite{animaltrack} - \textbf{Method - Total Instances:428753} &  \textbf{Precision}$\uparrow$&  \textbf{Recall\%}$\uparrow$ &  \textbf{TP}$\uparrow$ &  \textbf{FP}$\downarrow$& \textbf{FP\%}$\downarrow$ & \textbf{FN}$\downarrow$ &\textbf{FN\%}$\downarrow$ &\textbf{mAP50}$\uparrow$ &\textbf{mAP}$\uparrow$  & \textbf{F1}$\uparrow$ \\ \hline
    \textit{Fixed Box Selection / No SAM / No Positional Filter}  & 68.68 & 57.58 & 246872 & 112582& 26.26 & 181881 &42.42&55.65&22.52 &64.13 \\ 
    \textit{Fixed Box Selection / No SAM / Positional Filter}  & 69.38 & 58.85& 252255 & 111321 &25.96 & 176498&41.17 &58.19&24.19 &64.79 \\ 
    \textit{Variable Box Selection / No SAM / No Positional Filter}  & 88.75 & 77.57 & 332569 & 42151 &9.83 & 96184 &22.43& \textbf{83.88} & \textbf{55.17} &81.11 \\ 
    \textit{Variable Box Selection / No SAM / Positional Filter}  & \textbf{89.17} & 76.80 & 329288 & \textbf{39993} & \textbf{9.33} & 99465&23.20 &83.12&53.94 &80.78 \\ 
    \textit{Fixed Box Selection / SAM / No Positional Filter}  & 83.97 & 68.99 & 295808 & 56461 & 13.17 & 132945 & 31.01 & 75.59 & 39.64 & 75.00 \\ 
    \textit{Fixed Box Selection / SAM / Positional Filter}  & 84.75 & 68.90 & 295401 & 53158& 12.40 & 133352 &31.10&75.82&39.89 & 75.29 \\ 
    \textit{Variable Box Selection / SAM / No Positional Filter}  & 86.83 & \textbf{77.88} & \textbf{333922} & 50660 &11.82  & \textbf{94831} & \textbf{22.12} & 82.98 & 47.70 & 81.07 \\ 
    \textit{Variable Box Selection / SAM / Positional Filter}  & 87.01 & 77.52 & 332377 & 49632 & 11.58  & 96376 &  22.48 &82.79&47.52 & \textbf{81.21} \\ \hline
\end{tabular} }
\caption{Ablation Study - GMOT-40 and AnimalTrack. Best results for the experiments are in \textbf{bold}.}

\label{tab:yourtable}
\end{table*}

\subsection{Ablation Study}
\begin{figure}[h]
\includegraphics[width=8.5cm]{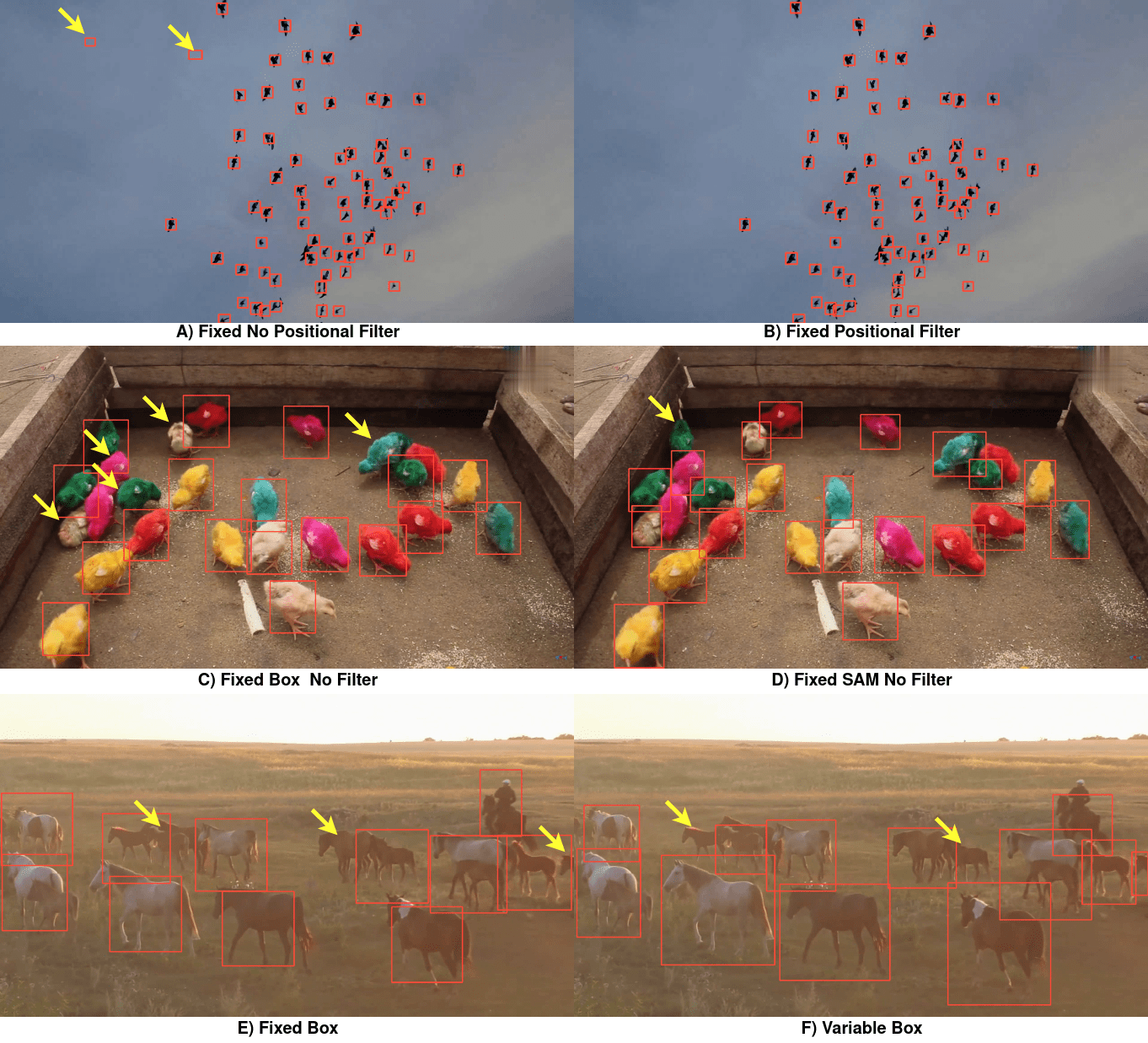}
 \caption{Comparison of pipeline components. Yellow arrows indicate changes in detection}
 \label{fig:ablationstudy}
\end{figure}
The goal of the ablation study is to validate the design choices of the POPCat pipeline. All results are collected  with the implementation detailed in section \ref{subsec:impl-details} for GMOT-40 and AnimalTrack. Visdrone was not included in the ablation study, as variables such as class distribution, variations in object sizing, ID switching, and scene entry cannot be ablated in an effective manner. 
\\

\subsubsection{Application of Positional Filtering}
\label{subsubsec:pos-filt}
Positional filtering as it relates to the pruning of objects, is used to control tracks that leave the extent of the image. This feature reduces the false positive percentage (FP\%) by eliminating erroneous annotations due to loss of track in the propagation stage.
The overall FP\% observes a reduction of 1.4\% (fixed box selection), 0.8\% (Variable box selection), 0.3\% (Fixed box with SAM), and 0.3\% (Variable box with SAM) for the GMOT-40 dataset. Positional filtering minimally impacts the recall\%, with increases under 1.0\%. 
For AnimalTrack, the overall FP\% is reduced by 0.3\% (fixed box selection), 0.5\% (variable box selection), 0.8\% (Fixed box with SAM), and 0.2\% (Variable box with SAM). The percentage recall increase by less than 1\% for AnimalTrack cases. Fig.\ref{fig:ablationstudy} depicts the effect of positional filtering on the output results. By removing lost tracks, we ensure the accuracy of the dataset used to train the detector. Therefore, the inference frames no longer show FPs as is seen between a) and b) Fig. \ref{fig:ablationstudy}. The characteristics of the object trajectories within a video dictate the impact of positional filtering on performance. Highly mobile targets will benefit more from positional filtering.

\subsubsection{Application of Segment Anything Masking} \label{subsubsec:SAMAPPS}
 The percent recall is increased by 13.3\% in GMOT-40, and 10.7\% in AnimalTrack between fixed selections and fixed SAM method. The use of SAM to improve the precision of bounding boxes halved the FP\% with a reduction of 12.4\% and 13.3\%  for the fixed box annotation of GMOT-40 and AnimalTrack. SAM has a lesser effect on variable box annotations with the recall increasing by 4.9\% for GMOT-40 and 0.5\% increase for AnimalTrack. For the variable box selection, the effect of SAM on FP\% is observed to be mixed, as GMOT-40 has a 0.8\% decrease, and AnimalTrack averages a 2.1\% increase. Elevated FP\% in AnimalTrack may be due to differences in the ground truth box size and detected size. A qualitative review of videos indicates that output box detections are smaller than ground-truth detections resulting in an incorrect FP assumption. Note the effectiveness of background exclusion in picture c) compared to d) in Fig.\ref{fig:ablationstudy}. Increases in recall are also observed in the higher number of detections in c) compared to d) Fig.\ref{fig:ablationstudy}. Based on these results, SAM is a critical addition to the POPCat pipeline. 
 
\subsubsection{Bounding Box Size Initialization}
\label{subsubsec:BBS}
Depending on the annotation requirements of the dataset, a fixed or variable box is required. GMOT-40 observes an average increase of 13.0\% in recall between fixed-box and variable-box methods. The SAM-based methods average an increase of 4.6\%. The FP\% is halved, an average decrease of 14.4\% from 23.7\% for fixed-box to variable-box, and 2.9\% from 8.4\% for the SAM-based methods is observed. Under AnimalTrack, the recall is increased by an average of 19.0\% (fixed box and variable box) and 8.7\% (fixed SAM and variable SAM). The FP\% is decreased by an average of 16.5\% from 26.0\% (fixed-box and variable-box) and 1.0\% from 12.8\% (fixed-SAM and variable-SAM). The performance improvements result from a better representation of the target data\cite{qualanno}. With a variable box selection, the target is fully enclosed in the annotation-sequence pair. This leads to a more accurate box representation of the target which improves the trained detector performance. This can be observed in Fig. \ref{fig:ablationstudy} between e) and f) where the detected area in f) fully encloses the targets when compared to e). The performance improvements are at the cost of labeling complexity and annotation time for the first frame. 

\section{Conclusion}
In this paper, we propose a novel annotation pipeline POPCat that utilizes point propagation and segmentation to train a detector for annotation of a given video. The detector results exceed the comparable literature review results for the GMOT-40, AnimalTrack, and Visdrone-2019 test data. Our pipeline exploits temporal consistencies in video sequences to propagate highly accurate labels with an overall manual-to-automatic annotation ratio of 1:274. To accommodate end-user preference, we propose a simple and complex method of pipeline initiation. We benchmark the results of such choices to illustrate improvements of up to 50\% false positive rates and recall. We report a variety of detection metrics to provide a complete picture of the performance for future comparisons. We ensure that the pipeline can operate on limited GPU hardware while maintaining a highly time-efficient annotation process. We achieve the goal of minimizing the human cost of annotation by limiting it to a subset of total frames. The pipeline achieves the goals of streamlining, simplifying, and improving the speed of custom video dataset creation.




%
{
    \small \bibliographystyle{IEEEtranBST/IEEEtran}
    \bibliography{IEEEtranBST/IEEEfull}

\begin{thebibliography}{10}
\providecommand{\url}[1]{#1}
\csname url@samestyle\endcsname
\providecommand{\newblock}{\relax}
\providecommand{\bibinfo}[2]{#2}
\providecommand{\BIBentrySTDinterwordspacing}{\spaceskip=0pt\relax}
\providecommand{\BIBentryALTinterwordstretchfactor}{4}
\providecommand{\BIBentryALTinterwordspacing}{\spaceskip=\fontdimen2\font plus
\BIBentryALTinterwordstretchfactor\fontdimen3\font minus \fontdimen4\font\relax}
\providecommand{\BIBforeignlanguage}[2]{{%
\expandafter\ifx\csname l@#1\endcsname\relax
\typeout{** WARNING: IEEEtran.bst: No hyphenation pattern has been}%
\typeout{** loaded for the language `#1'. Using the pattern for}%
\typeout{** the default language instead.}%
\else
\language=\csname l@#1\endcsname
\fi
#2}}
\providecommand{\BIBdecl}{\relax}
\BIBdecl

\bibitem{bai2021gmot}
H.~Bai, W.~Cheng, P.~Chu, J.~Liu, K.~Zhang, and H.~Ling, ``Gmot-40: A benchmark for generic multiple object tracking,'' in \emph{Proceedings of the IEEE/CVF Conference on Computer Vision and Pattern Recognition}, 2021, pp. 6719--6728.

\bibitem{animaltrack}
L.~Zhang, J.~Gao, Z.~Xiao, and H.~Fan, ``Animaltrack: A benchmark for multi-animal tracking in the wild,'' 2022.

\bibitem{Geiger2013IJRR}
A.~Geiger, P.~Lenz, C.~Stiller, and R.~Urtasun, ``Vision meets robotics: The kitti dataset,'' \emph{International Journal of Robotics Research (IJRR)}, 2013.

\bibitem{MOTS20}
\BIBentryALTinterwordspacing
P.~Voigtlaender, M.~Krause, A.~Osep, J.~Luiten, B.~B.~G. Sekar, A.~Geiger, and B.~Leibe, ``Mots: Multi-object tracking and segmentation,'' \emph{arXiv:1902.03604[cs]}, 2019, arXiv: 1902.03604. [Online]. Available: \url{http://arxiv.org/abs/1902.03604}
\BIBentrySTDinterwordspacing

\bibitem{Yolo}
\BIBentryALTinterwordspacing
G.~Jocher, A.~Chaurasia, and J.~Qiu, ``{YOLO by Ultralytics},'' Jan. 2023. [Online]. Available: \url{https://github.com/ultralytics/ultralytics}
\BIBentrySTDinterwordspacing

\bibitem{RCNN}
R.~Girshick, J.~Donahue, T.~Darrell, and J.~Malik, ``Rich feature hierarchies for accurate object detection and semantic segmentation,'' 2014.

\bibitem{carion2020end}
N.~Carion, F.~Massa, G.~Synnaeve, N.~Usunier, A.~Kirillov, and S.~Zagoruyko, ``End-to-end object detection with transformers,'' in \emph{European conference on computer vision}.\hskip 1em plus 0.5em minus 0.4em\relax Springer, 2020, pp. 213--229.

\bibitem{kirillov2023segany}
A.~Kirillov, E.~Mintun, N.~Ravi, H.~Mao, C.~Rolland, L.~Gustafson, T.~Xiao, S.~Whitehead, A.~C. Berg, W.-Y. Lo, P.~Doll{\'a}r, and R.~Girshick, ``Segment anything,'' \emph{arXiv:2304.02643}, 2023.

\bibitem{9021972}
P.~Zhu, D.~Du, L.~Wen, X.~Bian, H.~Ling, Q.~Hu, T.~Peng, J.~Zheng, X.~Wang, Y.~Zhang, L.~Bo, H.~Shi, R.~Zhu, B.~Dong, D.~R. Pailla, F.~Ni, G.~Gao, G.~Liu, H.~Xiong, J.~Ge, J.~Zhou, J.~Hu, L.~Sun, L.~Chen, M.~Lauer, Q.~Liu, S.~S. Chennamsetty, T.~Sun, T.~Wu, V.~A. Kollerathu, W.~Tian, W.~Qin, X.~Chen, X.~Zhao, Y.~Lian, Y.~Wu, Y.~Li, Y.~Li, Y.~Wang, Y.~Song, Y.~Yao, Y.~Zhang, Z.~Pi, Z.~Chen, Z.~Xu, Z.~Xiao, Z.~Luo, and Z.~Liu, ``Visdrone-vid2019: The vision meets drone object detection in video challenge results,'' in \emph{2019 IEEE/CVF International Conference on Computer Vision Workshop (ICCVW)}, 2019, pp. 227--235.

\bibitem{MOSP}
J.~Castaño-Amorós, F.~Fuentes, and P.~Gil, ``Mosppa: Monitoring system for palletised packaging recognition and tracking,'' \emph{The International Journal of Advanced Manufacturing Technology}, vol. 125, 02 2023.

\bibitem{rutinowski2023semi}
J.~Rutinowski, H.~Youssef, S.~Franke, I.~F. Priyanta, F.~Polachowski, M.~Roidl, and C.~Reining, ``Semi-automated computer vision based tracking of multiple industrial entities--a framework and dataset creation approach,'' \emph{arXiv preprint arXiv:2304.00950}, 2023.

\bibitem{rasmussen2022challenge}
C.~B. Rasmussen, K.~Kirk, and T.~B. Moeslund, ``The challenge of data annotation in deep learning—a case study on whole plant corn silage,'' \emph{Sensors}, vol.~22, no.~4, p. 1596, 2022.

\bibitem{dorr2020fully}
L.~D{\"o}rr, F.~Brandt, M.~Pouls, and A.~Naumann, ``Fully-automated packaging structure recognition in logistics environments,'' in \emph{2020 25th IEEE International Conference on Emerging Technologies and Factory Automation (ETFA)}, vol.~1.\hskip 1em plus 0.5em minus 0.4em\relax IEEE, 2020, pp. 526--533.

\bibitem{PALLETFORK}
R.-J. Hong, Y.-R. Li, M.-H. Hung, J.-W. Chang, and J.~C. Hung, ``Integrating object detection and semantic segmentation into automated pallet forking and picking system in agv,'' in \emph{Frontier Computing}, J.~C. Hung, N.~Y. Yen, and J.-W. Chang, Eds.\hskip 1em plus 0.5em minus 0.4em\relax Singapore: Springer Nature Singapore, 2022, pp. 121--129.

\bibitem{9613245}
Y.~Li, Y.~Niu, Y.~Liu, L.~Zheng, Z.~Wang, and W.~Zhe, ``Computer vision based conveyor belt congestion recognition in logistics industrial parks,'' in \emph{2021 26th IEEE International Conference on Emerging Technologies and Factory Automation (ETFA )}, 2021, pp. 1--8.

\bibitem{COAL}
H.~Pan, Y.~Shi, X.~Lei, Z.~Wang, and F.~Xin, ``Fast identification model for coal and gangue based on the improved tiny yolo v3,'' \emph{Journal of Real-Time Image Processing}, vol.~19, pp. 1--15, 06 2022.

\bibitem{9204369}
A.~Karami, M.~Crawford, and E.~J. Delp, ``Automatic plant counting and location based on a few-shot learning technique,'' \emph{IEEE Journal of Selected Topics in Applied Earth Observations and Remote Sensing}, vol.~13, pp. 5872--5886, 2020.

\bibitem{devlin2018bert}
J.~Devlin, M.-W. Chang, K.~Lee, and K.~Toutanova, ``Bert: Pre-training of deep bidirectional transformers for language understanding,'' \emph{arXiv preprint arXiv:1810.04805}, 2018.

\bibitem{Radford2018ImprovingLU}
\BIBentryALTinterwordspacing
A.~Radford and K.~Narasimhan, ``Improving language understanding by generative pre-training,'' 2018. [Online]. Available: \url{https://api.semanticscholar.org/CorpusID:49313245}
\BIBentrySTDinterwordspacing

\bibitem{tran2023z}
K.~H. Tran, T.-P. Nguyen, A.~D.~L. Dinh, P.~Nguyen, T.~Phan, K.~Luu, D.~Adjeroh, and N.~H. Le, ``Z-gmot: Zero-shot generic multiple object tracking,'' \emph{arXiv preprint arXiv:2305.17648}, 2023.

\bibitem{li2022grounded}
L.~H. Li, P.~Zhang, H.~Zhang, J.~Yang, C.~Li, Y.~Zhong, L.~Wang, L.~Yuan, L.~Zhang, J.-N. Hwang \emph{et~al.}, ``Grounded language-image pre-training,'' in \emph{Proceedings of the IEEE/CVF Conference on Computer Vision and Pattern Recognition}, 2022, pp. 10\,965--10\,975.

\bibitem{nguyen2022few}
T.~Nguyen, C.~Pham, K.~Nguyen, and M.~Hoai, ``Few-shot object counting and detection,'' in \emph{European Conference on Computer Vision}.\hskip 1em plus 0.5em minus 0.4em\relax Springer, 2022, pp. 348--365.

\bibitem{SEMSEG}
\BIBentryALTinterwordspacing
X.~Shi, S.~Zhang, M.~Cheng, L.~He, X.~Tang, and Z.~Cui, ``Few-shot semantic segmentation for industrial defect recognition,'' \emph{Computers in Industry}, vol. 148, p. 103901, 2023. [Online]. Available: \url{https://www.sciencedirect.com/science/article/pii/S0166361523000519}
\BIBentrySTDinterwordspacing

\bibitem{CVAT}
\BIBentryALTinterwordspacing
{CVAT.ai Corporation}, ``{Computer Vision Annotation Tool (CVAT)},'' Sep. 2022. [Online]. Available: \url{https://github.com/opencv/cvat}
\BIBentrySTDinterwordspacing

\bibitem{harley2022particle}
A.~W. Harley, Z.~Fang, and K.~Fragkiadaki, ``Particle video revisited: Tracking through occlusions using point trajectories,'' in \emph{European Conference on Computer Vision}.\hskip 1em plus 0.5em minus 0.4em\relax Springer, 2022, pp. 59--75.

\bibitem{ji2023segment}
W.~Ji, J.~Li, Q.~Bi, W.~Li, and L.~Cheng, ``Segment anything is not always perfect: An investigation of sam on different real-world applications,'' \emph{arXiv preprint arXiv:2304.05750}, 2023.

\bibitem{grabcut}
\BIBentryALTinterwordspacing
C.~Rother, V.~Kolmogorov, and A.~Blake, ``"grabcut": interactive foreground extraction using iterated graph cuts.'' \emph{ACM Trans. Graph.}, vol.~23, no.~3, pp. 309--314, 2004. [Online]. Available: \url{http://dblp.uni-trier.de/db/journals/tog/tog23.html#RotherKB04}
\BIBentrySTDinterwordspacing

\bibitem{autofit}
M.~Cruz, J.~Keh, M.~Rivera, N.~Velasco, J.~A. Jose, E.~Sybingco, E.~Dadios, W.~Madria, and A.~Miguel, ``Auto-fit: A human-machine collaboration feature for fitting bounding box annotations,'' in \emph{2020 IEEE 12th International Conference on Humanoid, Nanotechnology, Information Technology, Communication and Control, Environment, and Management (HNICEM)}, 2020, pp. 1--6.

\bibitem{8100272}
E.~Real, J.~Shlens, S.~Mazzocchi, X.~Pan, and V.~Vanhoucke, ``Youtube-boundingboxes: A large high-precision human-annotated data set for object detection in video,'' in \emph{2017 IEEE Conference on Computer Vision and Pattern Recognition (CVPR)}, 2017, pp. 7464--7473.

\bibitem{9573394}
P.~Zhu, L.~Wen, D.~Du, X.~Bian, H.~Fan, Q.~Hu, and H.~Ling, ``Detection and tracking meet drones challenge,'' \emph{IEEE Transactions on Pattern Analysis and Machine Intelligence}, vol.~44, no.~11, pp. 7380--7399, 2022.

\bibitem{Fujitake_2022_WACV}
M.~Fujitake and A.~Sugimoto, ``Video representation learning through prediction for online object detection,'' in \emph{Proceedings of the IEEE/CVF Winter Conference on Applications of Computer Vision (WACV) Workshops}, January 2022, pp. 530--539.

\bibitem{ristani2016performance}
E.~Ristani, F.~Solera, R.~Zou, R.~Cucchiara, and C.~Tomasi, ``Performance measures and a data set for multi-target, multi-camera tracking,'' in \emph{European conference on computer vision}.\hskip 1em plus 0.5em minus 0.4em\relax Springer, 2016, pp. 17--35.

\bibitem{electronics10030279}
\BIBentryALTinterwordspacing
R.~Padilla, W.~L. Passos, T.~L.~B. Dias, S.~L. Netto, and E.~A.~B. da~Silva, ``A comparative analysis of object detection metrics with a companion open-source toolkit,'' \emph{Electronics}, vol.~10, no.~3, 2021. [Online]. Available: \url{https://www.mdpi.com/2079-9292/10/3/279}
\BIBentrySTDinterwordspacing

\bibitem{GlobTrack}
L.~Huang, X.~Zhao, and K.~Huang, ``Globaltrack: A simple and strong baseline for long-term tracking,'' in \emph{Proceedings of the AAAI Conference on Artificial Intelligence}, vol.~34, no.~07, 2020, pp. 11\,037--11\,044.

\bibitem{10.1007/978-3-319-10602-1_48}
T.-Y. Lin, M.~Maire, S.~Belongie, J.~Hays, P.~Perona, D.~Ramanan, P.~Doll{\'a}r, and C.~L. Zitnick, ``Microsoft coco: Common objects in context,'' in \emph{Computer Vision -- ECCV 2014}, D.~Fleet, T.~Pajdla, B.~Schiele, and T.~Tuytelaars, Eds.\hskip 1em plus 0.5em minus 0.4em\relax Cham: Springer International Publishing, 2014, pp. 740--755.

\bibitem{liu2023siamesedetr}
Q.~Liu, Y.~Li, Y.~Jiang, and Y.~Fu, ``Siamese-detr for generic multi-object tracking,'' 2023.

\bibitem{Shao_2019_ICCV}
S.~Shao, Z.~Li, T.~Zhang, C.~Peng, G.~Yu, X.~Zhang, J.~Li, and J.~Sun, ``Objects365: A large-scale, high-quality dataset for object detection,'' in \emph{Proceedings of the IEEE/CVF International Conference on Computer Vision (ICCV)}, October 2019.

\bibitem{ren2016faster}
S.~Ren, K.~He, R.~Girshick, and J.~Sun, ``Faster r-cnn: Towards real-time object detection with region proposal networks,'' 2016.

\bibitem{law2020cornernet}
H.~Law and J.~Deng, ``Cornernet: Detecting objects as paired keypoints.'' \emph{International Journal of Computer Vision}, vol. 128, no.~3, 2020.

\bibitem{9010985}
K.~Duan, S.~Bai, L.~Xie, H.~Qi, Q.~Huang, and Q.~Tian, ``Centernet: Keypoint triplets for object detection,'' in \emph{2019 IEEE/CVF International Conference on Computer Vision (ICCV)}, 2019, pp. 6568--6577.

\bibitem{Lin_2017_CVPR}
T.-Y. Lin, P.~Dollar, R.~Girshick, K.~He, B.~Hariharan, and S.~Belongie, ``Feature pyramid networks for object detection,'' in \emph{Proceedings of the IEEE Conference on Computer Vision and Pattern Recognition (CVPR)}, July 2017.

\bibitem{feichtenhofer2017detect}
C.~Feichtenhofer, A.~Pinz, and A.~Zisserman, ``Detect to track and track to detect,'' in \emph{Proceedings of the IEEE international conference on computer vision}, 2017, pp. 3038--3046.

\bibitem{zhu2017flow}
X.~Zhu, Y.~Wang, J.~Dai, L.~Yuan, and Y.~Wei, ``Flow-guided feature aggregation for video object detection,'' in \emph{Proceedings of the IEEE international conference on computer vision}, 2017, pp. 408--417.

\bibitem{tian2019fcos}
Z.~Tian, C.~Shen, H.~Chen, and T.~He, ``Fcos: Fully convolutional one-stage object detection,'' 2019.

\bibitem{yolodat}
``mshamrai/yolov8x-visdrone,'' \url{https://huggingface.co/mshamrai/yolov8x-visdrone}, accessed: 2023-08-04.

\bibitem{qualanno}
C.~Agnew, C.~Eising, P.~Denny, A.~Scanlan, P.~Van De~Ven, and E.~M. Grua, ``Quantifying the effects of ground truth annotation quality on object detection and instance segmentation performance,'' \emph{IEEE Access}, vol.~11, pp. 25\,174--25\,188, 2023.

\end{thebibliography}
}

\end{document}